\documentclass[10pt,twocolumn,letterpaper]{article}
\usepackage{cvpr}              
\usepackage{subcaption} 
\definecolor{cvprblue}{rgb}{0.21,0.49,0.74}
\usepackage[breaklinks,colorlinks,allcolors=cvprblue]{hyperref}
\usepackage{url}

\def\paperID{41090} 
\def\confName{CVPR}
\def\confYear{2026}

\title{SATTC: Structure-Aware Label-Free Test-Time Calibration for Cross-Subject EEG-to-Image Retrieval}

\author{Qunjie Huang\\
Yunnan University, China\\
{\small \texttt{huangqunjie@stu.ynu.edu.cn}}
\and
Weina Zhu\thanks{Corresponding author.}\\
Yunnan University, China\\
{\small \texttt{zhuweina@ynu.edu.cn}}
}

\begin{document}
\maketitle
\begin{abstract}
Cross-subject EEG-to-image retrieval for visual decoding is challenged by subject shift and hubness in the embedding space, which distort similarity geometry and destabilize top-$k$ rankings, making small-$k$ shortlists unreliable. We introduce SATTC (Structure-Aware Test-Time Calibration), a label-free calibration head that operates directly on the similarity matrix of frozen EEG and image encoders. SATTC combines a geometric expert—subject-adaptive whitening of EEG embeddings with an adaptive variant of Cross-domain Similarity Local Scaling (CSLS)—and a structural expert built from mutual nearest neighbors, bidirectional top-$k$ ranks, and class popularity, fused via a simple Product-of-Experts rule. On THINGS-EEG2 under a strict leave-one-subject-out protocol, standardized inference with cosine similarities, $\ell_2$-normalized embeddings, and candidate whitening already yields a strong cross-subject baseline over the original ATM retrieval setup. Building on this baseline, SATTC further improves Top-1 and Top-5 accuracy, reduces hubness and per-class imbalance, and produces more reliable small-$k$ shortlists. These gains transfer across multiple EEG encoders, supporting SATTC as an encoder-agnostic, label-free test-time calibration layer for cross-subject neural decoding. Code is available at \url{https://github.com/QunjieHuang/SATTC-CVPR2026}

\vspace{-2mm} 
\end{abstract}

\section{Introduction} 
\label{sec:intro}

Decoding visual perception from brain activity is a long-standing goal in neuroscience and machine learning. Non-invasive electroencephalography (EEG) is particularly attractive because it is portable, relatively low-cost, and offers millisecond temporal resolution. Building on these advantages, recent work has demonstrated EEG-based visual classification, zero-shot EEG-to-image retrieval, and image reconstruction with diffusion models~\cite{spampinato2017deep,song2024decoding,li2024atm}. In EEG-to-image retrieval, an EEG signal is mapped into a joint embedding space and used to retrieve its nearest image neighbors. With the growing availability of large-scale EEG-image benchmarks and powerful cross-modal encoders~\cite{radford2021clip}, EEG-to-image retrieval pipelines can now be evaluated at scale, yet a central question emerges: how can we reliably retrieve rich visual semantics from non-invasive EEG signals in realistic cross-subject settings?

\begin{figure}[t]  
  \centering
  \includegraphics[width=\linewidth]{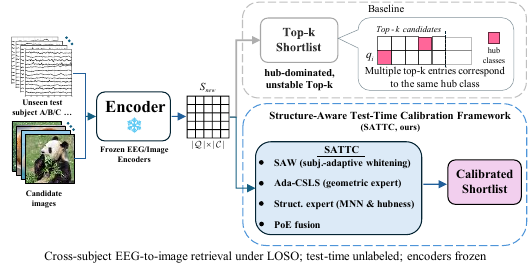} 
  \caption{Overview of cross-subject EEG-to-image retrieval under leave-one-subject-out (LOSO) evaluation and the proposed SATTC (Structure-Aware Test-Time Calibration) head. Frozen EEG and image encoders produce a similarity matrix whose baseline top-$k$ shortlist is hub-dominated; SATTC performs subject-adaptive whitening, adaptive CSLS, and a structural expert with product-of-experts (PoE) fusion to yield calibrated top-$k$ candidates in a label-free test-time setting.}
  \label{fig:intro_illustration}
  \vspace{-5mm} 
\end{figure}

Early work on EEG-based visual retrieval largely relied on supervised learning, training subject-specific classifiers or regressors on labeled EEG-image pairs to predict stimulus labels or features. However, such pipelines require costly trial-level annotations, are constrained by limited data, and typically generalize poorly to unseen subjects. As real cross-subject deployments seldom provide labels for new users, unsupervised and zero-shot paradigms have emerged as natural alternatives. Several recent studies adopt encoder-centric unsupervised or zero-shot EEG-to-image paradigms that align EEG and image embeddings via contrastive learning~\cite{song2024decoding,chen2024mind,song2025recognizing,li2024atm,chen2024necomimi,zhang2025category}. These works introduce increasingly expressive EEG encoders and report strong zero-shot recognition and retrieval performance on THINGS-EEG2 benchmarks~\cite{gifford2022large}, but their test-time evaluation typically reduces to non-standardized nearest-neighbor search over learned similarity scores in the shared space. In parallel, a separate line of EEG decoding research improves cross-subject robustness by learning subject-invariant or domain-adapted representations through adversarial inference, multi-source marginal distribution adaptation, and plug-and-play domain adaptation modules~\cite{ozdenizci2020learning,zhao2021plug,chen2021ms,she2023multisource}. While effective for representation learning, these methods operate at training time and have not been instantiated as generic, label-free calibration heads~\cite{sun2020ttt,zhang2022memo} that can standardize and improve retrieval shortlists in zero-shot cross-subject EEG-to-image settings.

Despite these advances, current unsupervised and zero-shot EEG-to-image retrieval pipelines still suffer from three test-time limitations: \textbf{(1) \emph{Missing structure-aware, label-free test-time calibration}.} Existing work largely focuses on encoder design while reducing inference to bare nearest-neighbor search, so the joint effect of cross-subject shift and hubness on top-$k$ rankings has not been systematically characterized, nor corrected, at test time. \textbf{(2) \emph{Lack of subject-adaptive, density-aware hubness mitigation}.} Most pipelines operate in a single globally normalized similarity space with simple $\ell_2$ or global whitening, or omit hubness correction altogether. When CSLS variants are used, they typically rely on a globally fixed neighborhood size tuned on labeled validation data, which fails to capture query- and class-specific local densities in strict zero-shot, cross-subject scenarios. \textbf{(3) \emph{Underused structural cues and absent shortlist calibration}.} Current pipelines treat each query in isolation and mainly report global Top-1/Top-5 or mAP, without leveraging mutual nearest neighbors, bidirectional top-$k$ ranks, or class popularity patterns to diagnose and correct small-$k$ shortlist quality and per-class imbalance. They also lack a reusable test-time head that can be attached to arbitrary cross-subject EEG-to-image encoders to improve Top-1 while preserving Top-5 and mitigating hubness.

To address these limitations, we propose SATTC (Structure-Aware Test-Time Calibration), a label-free calibration head that operates purely on the test-time similarity matrix of frozen EEG and image encoders. We first standardize the inference pipeline with cosine similarities, $\ell_2$-normalized embeddings, and candidate whitening, then combine a geometric expert—subject-adaptive whitening of EEG embeddings followed by an adaptive CSLS scheme based on row- and column-wise local densities—with a structural expert that leverages mutual nearest neighbors, bidirectional top-$k$ ranks, and class popularity patterns. The two experts are fused via a simple Product-of-Experts rule to recalibrate EEG–image similarity scores, explicitly mitigating hubness and producing more stable small-$k$ shortlists for downstream zero-shot decoding. Our main contributions are summarized as follows:

\begin{itemize}
\item We formulate cross-subject EEG-to-image retrieval as a structure-aware, label-free test-time calibration problem, and empirically characterize how subject shift and hubness jointly destabilize top-$k$ rankings in this setting.
\item We introduce a geometric expert that combines subject-adaptive whitening of EEG embeddings with an adaptive CSLS scheme, estimating row- and column-wise local densities to derive query- and class-dependent neighborhoods and reduce hubness without global-$k$ tuning.
\item We propose a structural expert that operates on the pre-CSLS similarity matrix using mutual nearest neighbors, bidirectional top-$k$ ranks, and class popularity, and fuse it with the geometric expert via a Product-of-Experts model. Experiments on the THINGS-EEG2 cross-subject benchmark show that our standardized inference baseline already outperforms the original ATM pipeline, and that SATTC further improves Top-1 accuracy over strong fixed-$k$ and adaptive CSLS baselines without harming Top-5 performance, while reducing hubness and per-class imbalance and yielding more reliable small-$k$ shortlists.

\end{itemize}

\section{Related Work}
\label{sec:related_work}

\subsection{EEG-based Visual Decoding and Cross-Subject Generalization}

EEG-based visual decoding has progressed from early coarse category recognition to fine-grained visual classification, image retrieval, and reconstruction~\cite{spampinato2017deep,li2024atm,song2025recognizing}. A common recipe is to learn an encoder that maps short EEG segments into a visual or multimodal embedding space, followed by nearest-neighbor retrieval or regression to image features on datasets such as THINGS-EEG2. Despite strong results, most systems are still evaluated in subject-specific or lightly adapted settings, assuming a separate model or labeled data for each target user, which underestimates the difficulty of deployment to unseen users without per-subject labels.

To improve cross-subject robustness, recent work explores subject-independent training, domain-invariant EEG features, and shared latent spaces using adversarial inference, multi-source marginal distribution adaptation, or plug-and-play domain adaptation modules~\cite{ozdenizci2020learning,chen2021ms,zhao2021plug,she2023multisource}. However, these approaches primarily focus on encoder architectures and training losses. They rarely analyze how similarity scores behave at test time under subject shift and hubness, nor how to recalibrate the retrieval geometry once the encoder is fixed. In contrast, we adopt a cross-subject EEG-to-image retrieval setup with frozen encoders and no target labels, and focus on label-free test-time calibration of the EEG-image similarity structure, complementary to representation-learning methods.

\subsection{Label-Free Test-Time Adaptation and Calibration}

Test-time adaptation and post-hoc calibration seek to reduce train-test mismatch by updating models or predictions using unlabeled test data~\cite{wang2020tent,niu2022efficient}. Many test-time adaptation (TTA) methods adapt batch-normalization statistics or minimize prediction entropy on streaming test samples, while post-hoc calibration approaches such as temperature scaling and its variants correct miscalibrated classifier confidences after training~\cite{guo2017calibration}. However, most of these techniques either require labeled validation data for tuning or implicitly rely on pseudo-labels and class-conditional statistics, which can be fragile for noisy, highly variable EEG data in cross-subject scenarios. Moreover, they are typically designed for closed-set classifier logits rather than dense similarity matrices arising in retrieval. 

We instead consider a strict label-free test-time regime where the EEG and image encoders are frozen and only the test-time similarity matrix is observable. Our goal is to design a calibration operator that acts directly on EEG--image similarities to stabilize top-$k$ rankings for downstream neural decoding.

\subsection{Hubness, CSLS, and Structure-Aware Retrieval Priors}

High-dimensional nearest-neighbor retrieval is strongly affected by hubness, where a few points appear in the top-$k$ lists of many queries and dominate rankings~\cite{dinu2015hubness,shigeto2015ridge}. Cross-domain Similarity Local Scaling (CSLS) mitigates hubness by using local neighborhoods to rescale similarity scores and down-weight globally ``popular'' candidates~\cite{conneau2017word}. Follow-up work refines CSLS or exploits mutual nearest neighbors and bidirectional ranks to discard inconsistent matches and sharpen alignments~\cite{zhong2017re,bai2017regularized,smith2017invertedsoftmax}, but typically assumes a fixed neighborhood size and treats hubness as a purely geometric effect. Rare classes in sparse regions can therefore be over-penalized, while dense hubs are not always fully suppressed when $k$ is small. 

Structure-aware re-ranking techniques based on graph diffusion~\cite{iscen2017diffusion}, mutual $k$NN graphs, or label propagation further exploit relationships among queries and candidates in image and cross-modal retrieval~\cite{bai2017regularized,zhong2017re}. Yet these methods usually rely on labeled data, transductive supervision, or iterative pseudo-labeling, and are not tailored to noisy cross-subject EEG where only small-$k$ shortlists and unlabeled test data are available. Our approach instead keeps the encoders fixed and operates purely on the pre-CSLS similarity matrix, combining an adaptive CSLS variant with query- and class-dependent neighborhoods and a structural prior built from mutual nearest neighbors, bidirectional top-$k$ ranks, and class popularity, fused via a Product-of-Experts formulation to obtain calibrated EEG--image retrieval scores.


\section{Method}
\label{sec:method}

\begin{figure*}[!t]
    \centering
    \includegraphics[width=0.96\textwidth]{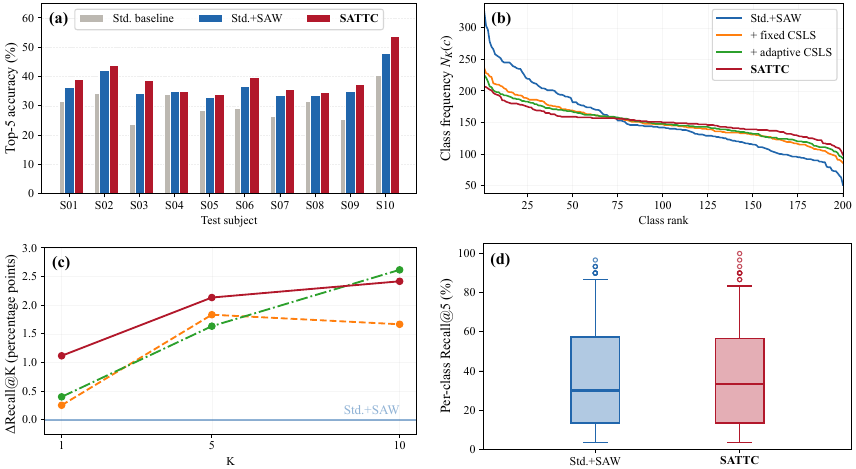}      
    \caption{\textbf{Effect of SAW and SATTC on subject shift, hubness, and shortlist quality.}
    (a) Per-subject Top-5 accuracy under LOSO.
    (b) Class popularity $N_K(c)$.
    (c) $\Delta$Recall@K over the Std.+SAW baseline.
    (d) Distribution of per-class Recall@5 for Std.+SAW and SATTC.
    SAW improves the standardized baseline, while SATTC further reduces hubness and yields more balanced and reliable small-K shortlists.}
    \label{fig:fig2}
    \vspace{-0.2em}
\end{figure*}






\subsection{Label-Free Test-Time Calibration: Problem Setup and Overview}

We consider cross-subject EEG-to-image retrieval on the THINGS-EEG2 dataset~\cite{Things2019image2,gifford2022large}, which provides multichannel EEG recordings while subjects view images from a fixed vocabulary of object categories. Let $Q$ denote EEG queries from unseen test subjects and $C$ the set of candidate image classes. A pretrained EEG encoder $f_{\text{eeg}}$ maps each query $q \in Q$ to an embedding $z_q \in \mathbb{R}^d$, and an image encoder $f_{\text{img}}$ maps each class $c \in C$ to an embedding $v_c \in \mathbb{R}^d$.
After subject-adaptive whitening and normalization (Section~\ref{sec:saw}), we obtain a pre-CSLS similarity matrix
\begin{equation}
S_{\mathrm{new}}(q,c)=\psi(\tilde z_q,\tilde v_c)
\label{eq:snew}
\end{equation}
where $\tilde{z}_q$ and $\tilde{v}_c$ denote the normalized EEG and image embeddings, and $\psi$ denotes the pre-CSLS scoring function specified in Section~\ref{sec:saw}. Retrieval is performed by ranking candidates $c \in C$ in descending order of $S_{\text{new}}(q,c)$, and our main evaluation metric is top-$k$ retrieval accuracy (typically $k \in \{1,5\}$) over queries from unseen subjects.

We adopt a strict label-free test-time regime. The encoders $f_{\text{eeg}}$ and $f_{\text{img}}$ are frozen; no labels or pseudo-labels from test subjects are available, and no adaptation of network weights is allowed. The only object we are allowed to modify at test time is the similarity structure itself. Formally, we seek a calibration operator
\begin{equation}
F: S_{\text{new}} \mapsto S_{\text{final}}
\label{eq:calibration-operator}
\end{equation}
that uses only unlabeled test-time statistics to produce calibrated scores $S_{\text{final}}(q,c)$ whose induced rankings better reflect true EEG–image correspondences and reduce hubness, yielding more reliable small-$k$ shortlists for downstream decoding.

We instantiate $F$ as a Product-of-Experts (PoE) fusion between two complementary experts. A \emph{geometric expert} $S_{\text{geom}}$ applies an adaptive CSLS scheme to $S_{\text{new}}$, using query- and class-dependent neighborhood sizes derived from local densities to mitigate hubness without tuning a global $k$ (Section~\ref{sec:ada_csls}). 
In parallel, a \emph{structural expert} $S_{\text{struct}}$ is built directly on the pre-CSLS similarities, exploiting mutual nearest neighbors, bidirectional top-$k$ ranks, and class popularity patterns as structure-aware priors (Section~\ref{sec:struct_expert}). Section~\ref{sec:poe} describes how these experts are fused via a simple PoE to obtain the final calibrated scores $S_{\text{final}}$. 

Fig.~\ref{fig:intro_illustration} illustrates this label-free test-time setting and the SATTC calibration head acting on the similarity matrix.

\subsection{Geometric Normalization via Subject-Adaptive Whitening}
\label{sec:saw}
Raw EEG embeddings exhibit strong subject-dependent shifts: even when different subjects perform the same task, their feature distributions can differ markedly in mean, variance, and covariance structure, leading to pronounced statistical and covariance shifts across subjects and sessions~\cite{albuquerque2019cross,ma2022large}.

We apply subject-adaptive whitening (SAW) to EEG features before computing similarities. Let $s(q)$ denote the subject index of query $q$, and
\begin{equation}
z_q = f_{\text{eeg}}(q) \in \mathbb{R}^d
\end{equation}
be the encoder output. For each subject $s$, we estimate a mean $\mu_s$ and covariance $\Sigma_s$ from the unlabeled embeddings in the current split: training embeddings for training subjects, and unlabeled test-time embeddings (or a calibration window) for the held-out subject. Construct a regularized whitening transform
\begin{equation}
W_s = (\Sigma_s + \lambda I)^{-\frac{1}{2}}
\end{equation}
with $\lambda > 0$ for numerical stability. Each query is first whitened as
\begin{equation}
\hat{z}_q = W_{s(q)} \bigl( z_q - \mu_{s(q)} \bigr)
\end{equation}
and then $\ell_2$-normalized to obtain
\begin{equation}
\tilde{z}_q = \frac{\hat{z}_q}{\|\hat{z}_q\|_2}
\end{equation}
This yields approximately zero-mean, identity-covariance, unit-norm EEG embeddings per subject and maps different subjects onto a shared sphere while preserving relative directions. We apply the same SAW transform during training and evaluation to maintain a consistent feature space.

On the image side, we optionally apply global whitening to candidate embeddings. Let
\begin{equation}
v_c = f_{\text{img}}(c)
\end{equation}
denote the candidate-side visual representation for class $c \in C$. We estimate a global mean $\mu_{\text{img}}$ and covariance $\Sigma_{\text{img}}$, form
\begin{equation}
\hat{v}_c = (\Sigma_{\text{img}} + \lambda I)^{-\frac{1}{2}} (v_c - \mu_{\text{img}})
\end{equation}
and normalize
\begin{equation}
\tilde{v}_c = \frac{\hat{v}_c}{\|\hat{v}_c\|_2}
\end{equation}
using $\tilde{v}_c$ as the normalized prototype; when whitening is disabled, we simply apply $\ell_2$-normalization to $v_c$.

Given normalized embeddings $\tilde{z}_q$ and $\tilde{v}_c$, we define a temperature-scaled cosine similarity
\begin{equation}
S_{\text{base}}(q, c) = \frac{\alpha}{\tau} \langle \tilde{z}_q, \tilde{v}_c \rangle
\label{eq:sbase}
\end{equation}
where $\alpha$ is a learned logit scale from contrastive pretraining and $\tau$ is a global temperature. Optionally, we apply global $z$-score normalization to $S_{\text{base}}$ using precomputed mean and variance over a held-out unlabeled set, yielding the pre-CSLS similarity matrix $S_{\text{new}}(q, c)$ used in subsequent stages.

\subsection{Adaptive CSLS Geometric Expert for Hubness Mitigation}
\label{sec:ada_csls}
Even after SAW-based normalization, the pre-CSLS similarity matrix $S_{\text{new}}(q, c)$ still exhibits hubness: a few image classes $c$ appear in the top-$k$ neighbors of many queries $q$, while rare classes are under-retrieved~\cite{radovanovic2010hubs,dinu2015hubness,lazaridou2015hubness}. Cross-domain Similarity Local Scaling (CSLS) is a standard remedy for hubness in cross-domain retrieval, using local-neighborhood averages to rescale similarity scores~\cite{conneau2017word}. Given similarities $s(q, c)$, CSLS defines
\begin{equation}
\mathrm{csls}(q, c) = 2\, s(q, c) - r_q(q) - r_c(c)
\label{eq:csls}
\end{equation}
where $r_q(q)$ and $r_c(c)$ are average similarities to the $k$ nearest neighbors of query $q$ and class $c$, respectively. By subtracting these local averages, CSLS down-weights candidates that are globally ``popular'' with many queries and partially compensates for local density.

However, standard CSLS uses a fixed neighborhood size $k$ for all queries and classes, implicitly assuming a roughly uniform density in the embedding space. Cross-subject EEG embeddings violate this assumption: some queries reside in sparse regions with only a few meaningful prototypes, whereas some classes form dense hubs that attract many unrelated queries. A global $k$ can thus under-penalize true hubs and over-penalize rare but correct neighbors, suppressing genuine matches from the top-$k$ list.

To address this, we construct an adaptive CSLS geometric expert that retains the CSLS form but derives query- and class-dependent neighborhoods from local densities on $S_{\text{new}}$. Let
\begin{equation}
s(q, c) = S_{\text{new}}(q, c)
\label{eq:snew-repeat}
\end{equation}
denote the normalized similarity between query $q$ and class $c$. For each query $q$, we estimate a row-wise local density
\begin{equation}
\rho_{\text{row}}(q) = \frac{1}{m} \sum_{c \in N^{(m)}_{\text{row}}(q)} s(q, c)
\label{eq:rho-row}
\end{equation}
where $N_{\text{row}}^{(m)}(q)$ is the set of $m$ most similar classes to $q$, with $m \gg K_{\max}$, the largest top-$K$ cutoff we evaluate (e.g.\ $K_{\max}=20$).
\begin{equation}
k_{\text{row}}(q) \in [k_{\min}, k_{\max}]
\label{eq:k-row}
\end{equation}
via a monotonic mapping from densities to integers in $[k_{\min}, k_{\max}]$, so that denser queries use larger neighborhoods and sparser queries use smaller ones.

Similarly, for each class $c$ we estimate a column-wise density $\rho_{\text{col}}(c)$, for example from how frequently $c$ appears in row-wise top-$K$ lists, and derive a class-specific neighborhood size
\begin{equation}
k_{\text{col}}(c) \in [k_{\min}, k_{\max}]
\label{eq:k-col}
\end{equation}
Given these adaptive neighborhood sizes, we compute
\begin{equation}
r_q(q) = \frac{1}{k_{\text{row}}(q)} \sum_{c' \in N^{(k_{\text{row}}(q))}_{\text{row}}(q)} s(q, c'),
\label{eq:rq}
\end{equation}
\begin{equation}
r_c(c) = \frac{1}{k_{\text{col}}(c)} \sum_{q' \in N^{(k_{\text{col}}(c))}_{\text{col}}(c)} s(q', c)
\label{eq:rc}
\end{equation}
where $N^{(k_{\text{row}}(q))}_{\text{row}}(q)$ and $N^{(k_{\text{col}}(c))}_{\text{col}}(c)$ denote the adaptive $k$-nearest neighbors along rows and columns.

Our adaptive CSLS score is then defined as
\begin{equation}
S_{\text{geom}}(q, c) = 2\, s(q, c) - r_q(q) - r_c(c)
\label{eq:sgeom}
\end{equation}
This construction preserves the familiar CSLS form while relaxing the fixed-density assumption: each query and class is normalized by a neighborhood size matched to its local density. All quantities are computed from $S_{\text{new}}$ alone, without access to any labels or pseudo-labels, and when the density mappings are constant $k_{\text{row}}(q) = k_{\text{col}}(c) = k_0$, our formulation reduces exactly to standard fixed-$k$ CSLS. We use $S_{\text{geom}}$ as the geometric expert in our Product-of-Experts fusion.

\subsection{Structural Expert from Pre-CSLS Similarity Evidence}
\label{sec:struct_expert}

While the adaptive CSLS geometric expert mitigates hubness from a local-density perspective, the pre-CSLS similarity matrix $S_{\text{new}}$ already contains structural patterns that correlate with reliable matches and hubs. We convert these patterns into a structural expert $S_{\text{struct}}$ that complements $S_{\text{geom}}$ without changing the encoder or recomputing similarities.

Given $S_{\text{new}}(q, c)$, we compute row- and column-wise ranks
\begin{equation}
\begin{aligned}
r_{\text{row}}(q,c)
  &= 1 + \#\{\, c' \mid S_{\text{new}}(q,c') > S_{\text{new}}(q,c) \,\},\\[-2pt]
r_{\text{col}}(c,q)
  &= 1 + \#\{\, q' \mid S_{\text{new}}(q',c) > S_{\text{new}}(q,c) \,\}
\end{aligned}
\label{eq:ranks}
\end{equation}
where exact score ties in $S_{\mathrm{new}}$ are broken deterministically. A pair $(q,c)$ is a strict mutual nearest neighbor (MNN@1) if
\begin{equation}
r_{\text{row}}(q,c) = r_{\text{col}}(c,q) = 1
\label{eq:mnn}
\end{equation}
and we also consider bidirectional top-$L$ pairs
\begin{equation}
r_{\text{row}}(q,c) \le L,\quad 
r_{\text{col}}(c,q) \le L
\label{eq:bidirectional-topL}
\end{equation}
which extend strict MNN@1 to slightly lower-confidence but still consistent matches.
We define a class-popularity statistic
\begin{equation}
\rho(c) = \#\{ q \mid r_{\text{row}}(q, c) \le K \}
\label{eq:popularity}
\end{equation}
the number of queries for which class $c$ appears in the row-wise top-$K$. Large $\rho(c)$ indicates hub-like behavior, whereas small $\rho(c)$ indicates rare or under-retrieved classes. Normalizing $\rho(c)$ across classes by min-max scaling yields a hubness score $h(c)\in[0,1]$ used to scale penalties on suspected hubs.

Using these quantities, we distinguish three structural cases over $(q, c)$:
\textbf{(i) \emph{anchors}}, the strict MNN@1 pairs in \eqref{eq:mnn};
\textbf{(ii) \emph{bidirectional top-$L$ pairs}}, satisfying \eqref{eq:bidirectional-topL};
\textbf{(iii) \emph{hub-like candidates}}, where $r_{\text{row}}(q, c) > K$, $r_{\text{col}}(c, q)$ is small, and $h(c)$ is large, i.e., $c$ acts as a hub across many queries but is not locally supported for this query.

We construct a structural logit matrix $S_{\text{struct}} \in \mathbb{R}^{|Q|\times|C|}$, initialized to zero, and update
\begin{equation}
S_{\text{struct}}(q,c) =
\begin{cases}
+\lambda_{\text{anchor}}, & (q,c) \in \mathcal{A}\\
-\lambda_{\text{pen}} \, h(c), & (q,c) \in \mathcal{H}\\
0, & \text{otherwise}
\end{cases}
\label{eq:sstruct}
\end{equation}
where $A \subset Q \times C$ collects anchors and strong bidirectional top-$L$ pairs, $H \subset Q \times C$ collects hub-like candidates, and $\lambda_{\text{anchor}}, \lambda_{\text{pen}} > 0$ control the strength of bonus and penalty. Penalties are masked on pairs in $A$, so high-confidence structural matches are only boosted, never suppressed. All decisions in \eqref{eq:sstruct} depend solely on $S_{\text{new}}$; $S_{\text{geom}}$ and the encoder are not used, and $S_{\text{new}}$ itself is never modified.

In summary, $S_{\text{struct}}$ provides a structure-aware prior over query–class pairs: it locks mutual top-1 and bidirectional top-$L$ matches via positive bias and down-weights ubiquitous hub classes, and, being computed once from pre-CSLS evidence and then held fixed, regularizes the final rankings without iterative self-reinforcement.

\subsection{Product-of-Experts Fusion and Final Scoring}
\label{sec:poe}

We combine the geometric expert $S_{\text{geom}}$ (adaptive CSLS on $S_{\text{new}}$) and the structural expert $S_{\text{struct}}$ (pre-CSLS evidence; Section~\ref{sec:struct_expert}) into a single calibrated scoring function. For each query $q$, the geometric expert defines an unnormalized distribution
\begin{equation}
p_{\text{geom}}(c \mid q) \propto \exp\bigl(S_{\text{geom}}(q, c)\bigr)
\label{eq:p-geom}
\end{equation}
and the structural expert defines
\begin{equation}
p_{\text{struct}}(c \mid q) \propto \exp\bigl(S_{\text{struct}}(q, c)\bigr)
\label{eq:p-struct}
\end{equation}
A simple Product-of-Experts (PoE) fusion then gives
\begin{equation}
p_{\text{final}}(c \mid q) \propto p_{\text{geom}}(c \mid q)^{\alpha}\, p_{\text{struct}}(c \mid q)^{\beta}
\label{eq:poe}
\end{equation}
where $\alpha,\beta \ge 0$ control the relative influence of the two experts. In logit space this reduces to a weighted sum
\begin{equation}
S_{\text{final}}(q, c) = \alpha\, S_{\text{geom}}(q, c) + \beta\, S_{\text{struct}}(q, c)
\label{eq:s-final}
\end{equation}

In practice we fix $\alpha = 1$ and tune a single scalar hyperparameter $\beta$, making the fusion lightweight and interpretable. Setting $\beta = 0$ recovers pure adaptive CSLS, while $\beta > 0$ progressively enforces the structure-aware prior encoded by $S_{\text{struct}}$: mutual top-1 and bidirectional top-$L$ matches receive a constant positive bias and hub-like candidates receive a negative bias proportional to their popularity level (Section~\ref{sec:struct_expert}). Because $S_{\text{struct}}$ is computed once from the pre-CSLS matrix $S_{\text{new}}$ and never updated based on $S_{\text{final}}$, the PoE acts as a single-shot regularizer rather than an iterative self-reinforcement scheme.

At test time, the label-free calibration operator from Section~3.1 can thus be written as
\begin{equation}
\mathcal{T}_{\beta}(S_{\text{new}}) := \mathrm{PoE}_{\beta}(S_{\text{geom}}, S_{\text{struct}}) = S_{\text{final}}
\label{eq:operator}
\end{equation}
where $S_{\text{geom}}$ and $S_{\text{struct}}$ are the geometric and structural experts applied to $S_{\text{new}}$. Top-$K$ retrieval is obtained by ranking classes $c \in C$ according to $S_{\text{final}}(q, c)$ for each query $q$.

\section{Experiments}
\label{sec:experiments}

\subsection{Experimental Setup}

\paragraph{Dataset and cross-subject protocol.}
We evaluate on the public THINGS-EEG2 dataset~\cite{Things2019image2,gifford2022large}, which provides EEG recordings from 10 subjects viewing images from 200 object categories. We adopt a leave-one-subject-out (LOSO) protocol: in each of 10 folds, one subject is held out for testing and the remaining nine are used for training and tuning.

To separate tuning from final evaluation, we use a nested LOSO procedure. From the nine training subjects, we select three representative dev subjects (easy/medium/hard) via a subject-level distance analysis and use only them for hyperparameter and epoch selection. We then fix the selected configuration, train a single model on all nine training subjects, and perform one-shot evaluation on the held-out subject. All hyperparameters were tuned once on a fixed dev pack from the training folds and then frozen across all LOSO folds and across encoders; at test time we perform label-free calibration using only unlabeled EEG from the held-out subject, and test labels are used solely for evaluation.

\paragraph{Evaluation metrics and baselines.}
We evaluate 200-way EEG-to-image retrieval: for each test EEG query, the model ranks all 200 candidate image classes. We report Top-5 accuracy as the primary metric and Top-1 accuracy as a secondary identification metric, and additionally show Recall@$K$ curves and hubness statistics in later sections.

To isolate the contribution of each component in our calibration pipeline, we compare a sequence of seven models, labeled ATM, Standardized baseline, +SAW, +SAW+CW, +CSLS, +Ada-CSLS, and SATTC (ours) in Table~\ref{tab:main_results}. Starting from the original ATM cross-subject baseline and our standardized baseline with cosine similarity, $\ell_2$-normalized features, and candidate-side whitening but no SAW or test-time calibration (“Standardized baseline” in Table~\ref{tab:main_results}), we incrementally add subject-adaptive whitening on EEG (+SAW), fixed-$k$ CSLS (+CSLS), adaptive CSLS as the geometric expert (+Ada-CSLS), and finally the structural PoE module on top of adaptive CSLS (SATTC). All variants share the same training setup: AdamW optimizer, batch size 1024, learning rate $5\times 10^{-4}$, and a fixed temperature $\tau=1.0$, trained and evaluated on a single NVIDIA RTX 4090 GPU.

\subsection{Cross-Subject EEG-to-Image Retrieval Results}

\paragraph{Overall comparison with baselines.}
Table~\ref{tab:main_results} reports 200-way cross-subject retrieval averaged over all LOSO folds and three seeds. Cosine similarity with $\ell_2$-normalized features and
candidate-side whitening (same encoder) already improves
substantially over the original ATM baseline with unnormalized
dot products, yielding the “Standardized baseline” row in Table~\ref{tab:main_results}.

Introducing subject-adaptive whitening on the EEG side (SAW) on top of the cosine + candidate-side whitening (CW) baseline brings a clear gain and serves as our cross-subject reference, showing that subject-specific normalization is crucial in the LOSO setting. On this SAW configuration (with candidate whitening), both fixed-$k$ CSLS and adaptive CSLS
further increase Top-5 accuracy while keeping Top-1 comparable; the adaptive variant matches the best fixed-$k$ configuration without a single global $k$, validating the geometric expert. Finally, adding the structural PoE module on top of adaptive CSLS (SATTC, ours) gives the strongest end-to-end result: relative to the SAW baseline, both Top-1 and Top-5 improve, while relative to adaptive CSLS, SATTC preserves Top-5 and improves Top-1 through structure-aware refinement.

\paragraph{Subject- and seed-wise stability.}
Across three random seeds per LOSO fold, standard deviations over subjects and seeds are small and the relative ordering between the baseline, SAW, CSLS/Ada-CSLS, and SATTC is consistent, indicating that our label-free test-time calibration is robust to initialization and subject splits.

\subsection{Ablations for Label-Free Test-Time Calibration}

\paragraph{Effect of SAW and candidate whitening.}

Table~\ref{tab:main_results} summarizes the whitening ablations. Compared to our standardized baseline (cosine + $\ell_2$ + CW), adding subject-adaptive whitening on the EEG side (+SAW) is the dominant source of improvement:
Top-5 rises from 30.5\% to 36.4\% and Top-1 from 9.2\% to 13.7\%. This shows that normalizing subject-specific EEG statistics is the key ingredient for cross-subject generalization under LOSO. Further adjusting candidate whitening in the presence of SAW (+SAW+CW) yields only a small additional Top-5 gain (36.8\%) with comparable Top-1, so the direct contribution of candidate-side whitening to retrieval remains modest once EEG features have been subject-normalized.

However, we retain candidate whitening (CW) primarily to support the structural
expert rather than chase additional accuracy. With SAW
alone, similarities are normalized row-wise, but column
scales can still vary across classes. Under SAW+CW, the
pre-CSLS similarity matrix $S_{\text{new}}$ used in SATTC
has both query-wise rows and class-wise columns on a
comparable scale, making class-popularity $N K(c)$, mutual
nearest neighbors, and other rank-based statistics derived
from $S_{\text{new}}$ (Section~\ref{sec:struct_expert}) more stable. We therefore
adopt SAW+CW as the default backbone for all subsequent
CSLS variants and SATTC experiments despite its modest
standalone gain over +SAW.

\begin{table}[t]
  \centering
  \small
  \setlength{\tabcolsep}{4pt}
  \caption{Cross-subject 200-way EEG-to-image retrieval results on THINGS-EEG2 under LOSO. Mean Top-5 and Top-1 accuracy (\%) over all test folds and three random seeds for the ATM baseline, our standardized baseline (cosine + $\ell_2$ normalization + candidate whitening), and successive additions of SAW, candidate whitening, CSLS variants, and SATTC.
  }
  \vspace{-0.5em}
  \label{tab:main_results}
  \begin{tabular}{lcc}
    \toprule
    Method & Top-5 (\%) $\uparrow$ & Top-1 (\%) $\uparrow$ \\
    \midrule
    ATM (original)              &  20.0 &  5.5 \\
    Standardized baseline       &  30.5 & 9.2 \\
    + SAW                       &  36.4 & 13.7 \\
    + SAW + CW                  &  36.8 & 13.5 \\
    + SAW + CW + CSLS           &  38.1 & 14.1 \\
    + SAW + CW + Ada-CSLS       &  \textbf{38.8} & 13.9 \\
    \textbf{SATTC (ours)}       &  38.4 & \textbf{14.8} \\
    \bottomrule
  \end{tabular}
  \vspace{-0.9em}
\end{table}

\medskip
\paragraph{Fixed-$k$ vs.\ adaptive CSLS.}
Table~\ref{tab:main_results} reports results for fixed-$k$ CSLS (with $k=12$ tuned on dev subjects) and our adaptive CSLS applied on the SAW+CW configuration. The two variants obtain very similar Top-1 and Top-5 accuracy, with
adaptive CSLS trading a small decrease in Top-1 for a slightly higher Top-5 while avoiding a globally chosen neighborhood size. Their hubness profiles, however, differ: as shown in Fig.~\ref{fig:fig2}, adaptive CSLS further flattens the class-occurrence curve and reduces the dominance of a few hub classes compared to fixed-$k$ CSLS, leading to a more balanced use of prototypes across ranks. This supports our choice of adaptive CSLS as the geometric expert: it preserves retrieval accuracy while providing stronger and more flexible hubness mitigation than a single global $k$.

\medskip
\noindent\textbf{Impact of structural expert and PoE.}
Finally, Table~\ref{tab:main_results} isolates the contribution of the structural expert and PoE by comparing +SAW+CW+Ada-CSLS with our full SATTC model (+Ada-CSLS+PoE). SATTC keeps Top-5 accuracy essentially unchanged (38.4 vs.\ 38.8\%) while providing a clear Top-1 gain (14.8 vs.\ 13.9\%), highlighting that the structural expert sharpens strict identification while preserving overall retrieval coverage. Consistent with this, Fig.~\ref{fig:fig2} shows that the structural PoE head in SATTC attains the largest $\Delta$Recall@K over the Std.+SAW baseline for small $K$ (1 and 5); relative to adaptive CSLS alone, it recovers more classes whose true label appears within small-K shortlists, reflecting better class-wise calibration even when the global Top-5 score remains comparable.

Practically, the calibration pipeline admits two deployment modes. For single-trial or online inference, the SAW+CSLS branch can be calibrated once from a small unlabeled window and then applied to subsequent trials with frozen parameters, whereas the structural expert and PoE serve as a batch refinement stage when multiple unlabeled test queries are available. In strict disjoint $N$-shot experiments, a one-time unlabeled window of $N=50$ already reaches 94.8\% of the $N=200$ SAW+CSLS upper bound. The PoE hyperparameter is also robust: a broad region of $\beta$ values yields near-identical Top-5 performance, and the default $\beta=1.9$ is within 0.1 percentage points of the best setting. Importantly, the full pipeline shows no end-to-end Top-1/Top-5 trade-off: from the SAW baseline to SATTC, Top-1 improves from 13.7\% to 14.8\%, while Top-5 improves from 36.4\% to 38.4\%.

\subsection{Hubness Reduction and Shortlist Quality}

\paragraph{Hubness and class popularity distribution.}
We use the class-popularity statistic $N_K(c)$ from Section~\ref{sec:struct_expert} as a hubness indicator (Fig.~\ref{fig:fig2}). Under the SAW+CW
configuration, fixed-$k$ CSLS shortens the heavy tail of $N_K(c)$
relative to the cosine baseline but still leaves some hubs; adaptive
CSLS further flattens $N_K(c)$, while SATTC yields the most uniform profile on this class-popularity statistic by suppressing spurious hubs and boosting underused classes.

\medskip
\paragraph{Per-class fairness and Recall@$K$.}
Per-class performance is evaluated with Recall@5 boxplots and Recall@$K$ curves for $K \in \{1,2,5,10,20\}$ (Fig.~\ref{fig:fig2}). Compared to the SAW+CW baseline, both fixed-$k$ CSLS and adaptive CSLS increase mean Recall@5; adaptive CSLS also achieves a higher median and lower variance across classes. SATTC further raises median Recall@5, narrows the spread, and shows the largest gains for small $K \le 5$, the regime most relevant for downstream decoding.

\begin{table}[t]
  \centering
  \small
  \setlength{\tabcolsep}{4pt}
  \caption{
    Plug-and-play generalization of SATTC across EEG encoders
    on 200-way THINGS-EEG2 cross-subject retrieval. For each
    encoder, we report Top-5 and Top-1 accuracy (\%) averaged
    over all LOSO folds and three random seeds, comparing our
    standardized inference baseline to the same encoder with
    SATTC applied as a label-free test-time calibration head.
  }
  \label{tab:plug_and_play}
  \begin{tabular}{lcccc}
    \toprule
    Encoder & \multicolumn{2}{c}{Top-5 $\uparrow$} & \multicolumn{2}{c}{Top-1 $\uparrow$} \\
    \cmidrule(lr){2-3} \cmidrule(lr){4-5}
            & Baseline & +SATTC & Baseline & +SATTC \\
    \midrule
    ATM                  &  30.5 & \textbf{38.4} &  9.2 & \textbf{14.8} \\
    EEGNetV4             &  20.5 & \textbf{34.8} &  5.4 & \textbf{10.8} \\
    EEGConformer         &  11.6 & \textbf{23.2} &  2.5 & \textbf{6.9} \\
    ShallowFBCSPNet      &  14.6 & \textbf{30.8} &  3.5 & \textbf{11.1} \\
    \bottomrule
  \end{tabular}
\end{table}

\subsection{Encoder-Agnostic Plug-and-Play Calibration}

We next test whether SATTC is tied to a specific EEG encoder or can act as a generic calibration layer. Table~\ref{tab:plug_and_play} reports 200-way cross-subject retrieval on THINGS-EEG2 for four heterogeneous encoders (ATM, EEGNetV4, EEGConformer, and ShallowFBCSPNet), spanning CSP-style, compact CNN, and transformer architectures. For each encoder we compare our standardized inference baseline to the same model equipped with SATTC as a label-free test-time calibration head, without changing the backbone or using labels from test subjects.

SATTC consistently improves all encoders. Top-5 accuracy increases by roughly $8$–$16$ percentage points and Top-1 by about $4$–$8$ points across backbones (e.g., from $30.5$ to $38.4$ Top-5 and $9.2$ to $14.8$ Top-1 for ATM, and from $20.5$ to $34.8$ and $5.4$ to $10.8$ for EEGNetV4), with similar gains for EEGConformer and ShallowFBCSPNet. These results indicate that SATTC operates on the geometry of the EEG–image similarity space rather than on encoder details, and can be plugged into existing cross-subject EEG pipelines as an encoder-agnostic, label-free test-time calibration module.


\section{Discussion and Conclusion}
\label{sec:discussion}

\noindent\textbf{Limitations.}
SATTC is currently evaluated only on THINGS-EEG2 under a LOSO cross-subject retrieval protocol. Its structural expert is hand-crafted from ranks, mutual nearest neighbors, and class popularity, and the current implementation operates on precomputed similarity matrices. In practice, the SAW+CSLS branch supports calibration-once then frozen inference, whereas the structural expert and PoE are most naturally applied in a batch refinement stage when multiple unlabeled test queries are available. Extending the method to additional datasets, encoder families, and decoding regimes such as retrieval-then-generation or EEG/fMRI reconstruction~\cite{li2024atm,chen2024necomimi} remains an important direction for future work.

\noindent\textbf{Conclusion.}
We cast cross-subject EEG-to-image retrieval as a label-free test-time calibration problem on similarity matrices. SATTC operates on frozen EEG and image encoders through a modular calibration head that combines subject-adaptive whitening, adaptive CSLS, and a lightweight structural prior from pre-CSLS similarities. On THINGS-EEG2 under strict LOSO evaluation, SATTC improves Top-1 and Top-5 over strong standardized baselines, reduces hubness and per-class imbalance, and yields more reliable small-$k$ shortlists. Together with its plug-and-play gains across multiple EEG encoders, these results suggest that similarity-space calibration is a practical route toward more robust cross-subject neural decoding.

{
    \small
    \bibliographystyle{ieeenat_fullname}
    \bibliography{main}

@String(CVPR= {IEEE Conf. Comput. Vis. Pattern Recog.})

@String(AAAI = {AAAI})

@String(CVPR  = {CVPR})

@inproceedings{song2024decoding,
  title = {Decoding {{Natural Images}} from {{EEG}} for {{Object Recognition}}},
  author = {Song, Yonghao and Liu, Bingchuan and Li, Xiang and Shi, Nanlin and Wang, Yijun and Gao, Xiaorong},
  booktitle = {International {{Conference}} on {{Learning Representations}}},
  year = {2024},
}

@article{song2025recognizing,
  title   = {Recognizing Natural Images From EEG With Language-Guided Contrastive Learning},
  author  = {Song, Yonghao and Wang, Yijun and He, Huiguang and Gao, Xiaorong},
  journal = {IEEE Transactions on Neural Networks and Learning Systems},
  volume  = {36},
  number  = {9},
  pages   = {15896--15910},
  year    = {2025},
  doi     = {10.1109/TNNLS.2025.3562743}
}

@inproceedings{li2024atm,
    author = {Li, Dongyang and Wei, Chen and Li, Shiying and Zou, Jiachen and Liu, Quanying},
     booktitle = {Advances in Neural Information Processing Systems},
     doi = {10.52202/079017-3266},
     editor = {A. Globerson and L. Mackey and D. Belgrave and A. Fan and U. Paquet and J. Tomczak and C. Zhang},
     pages = {102822--102864},
     publisher = {Curran Associates, Inc.},
     title = {Visual Decoding and Reconstruction via EEG Embeddings with Guided Diffusion},
     url = {https://proceedings.neurips.cc/paper_files/paper/2024/file/ba5f1233efa77787ff9ec015877dbd1f-Paper-Conference.pdf},
     volume = {37},
     year = {2024}
}

@article{chen2024mind,
  title={Mind's eye: image recognition by EEG via multimodal similarity-keeping contrastive learning},
  author={Chen, Chi-Sheng and Wei, Chun-Shu},
  journal={arXiv preprint arXiv:2406.16910},
  year={2024}
}

@article{chen2024necomimi,
  title={Necomimi: Neural-cognitive multimodal eeg-informed image generation with diffusion models},
  author={Chen, Chi-Sheng},
  journal={arXiv preprint arXiv:2410.00712},
  year={2024}
}

@inproceedings{zhang2025category,
  title={Category-aware EEG image generation based on wavelet transform and contrast semantic loss},
  author={Zhang, Enshang and Zhang, Zhicheng and Hanakawa, Takashi},
  booktitle = {Proceedings of the Thirty-Fourth International Joint Conference on Artificial Intelligence, {IJCAI-25}},
  publisher = {International Joint Conferences on Artificial Intelligence Organization},
  editor    = {James Kwok},
  pages     = {7922--7930},
  year      = {2025},
  month     = {8},
  note      = {Main Track},
  doi       = {10.24963/ijcai.2025/881},
  url       = {https://doi.org/10.24963/ijcai.2025/881},
}

@article{gifford2022large,
  title={A large and rich EEG dataset for modeling human visual object recognition},
  author={Gifford, Alessandro T and Dwivedi, Kshitij and Roig, Gemma and Cichy, Radoslaw M},
  journal={NeuroImage},
  volume={264},
  pages={119754},
  year={2022},
  publisher={Elsevier}
}

@article{ozdenizci2020learning,
  title={Learning invariant representations from EEG via adversarial inference},
  author={{\"O}zdenizci, Ozan and Wang, Ye and Koike-Akino, Toshiaki and Erdo{\u{g}}mu{\c{s}}, Deniz},
  journal={IEEE Access},
  volume={8},
  pages={27074--27085},
  year={2020},
  publisher={IEEE}
}

@inproceedings{zhao2021plug,
  title={Plug-and-play domain adaptation for cross-subject EEG-based emotion recognition},
  author={Zhao, Li-Ming and Yan, Xu and Lu, Bao-Liang},
  booktitle={Proceedings of the AAAI conference on artificial intelligence},
  volume={35},
  number={1},
  pages={863--870},
  year={2021}
}

@article{chen2021ms,
  title={MS-MDA: Multisource marginal distribution adaptation for cross-subject and cross-session EEG emotion recognition},
  author={Chen, Hao and Jin, Ming and Li, Zhunan and Fan, Cunhang and Li, Jinpeng and He, Huiguang},
  journal={Frontiers in Neuroscience},
  volume={15},
  pages={778488},
  year={2021},
  publisher={Frontiers Media SA}
}

@article{she2023multisource,
  title={Multisource associate domain adaptation for cross-subject and cross-session EEG emotion recognition},
  author={She, Qingshan and Zhang, Chenqi and Fang, Feng and Ma, Yuliang and Zhang, Yingchun},
  journal={IEEE Transactions on Instrumentation and Measurement},
  volume={72},
  pages={1--12},
  year={2023},
  publisher={IEEE}
}

@inproceedings{conneau2017word,
  title     = {Word Translation Without Parallel Data},
  author    = {Lample, Guillaume and Conneau, Alexis and Ranzato, Marc'Aurelio and Denoyer, Ludovic and J{\'e}gou, Herv{\'e}},
  booktitle = {International Conference on Learning Representations},
  year      = {2018},
  url       = {https://openreview.net/forum?id=H196sainb}
}

@inproceedings{zhong2017re,
  title={Re-ranking person re-identification with k-reciprocal encoding},
  author={Zhong, Zhun and Zheng, Liang and Cao, Donglin and Li, Shaozi},
  booktitle={Proceedings of the IEEE conference on computer vision and pattern recognition},
  pages={1318--1327},
  year={2017}
}

@inproceedings{bai2017regularized,
  author = {Bai, Song and Bai, Xiang and Tian, Qi and Latecki, Longin Jan},
title = {Regularized diffusion process for visual retrieval},
year = {2017},
publisher = {AAAI Press},
booktitle = {Proceedings of the Thirty-First AAAI Conference on Artificial Intelligence},
pages = {3967–3973},
numpages = {7},
location = {San Francisco, California, USA},
series = {AAAI'17}
}

@inproceedings{wang2020tent,
  title     = {Tent: Fully Test-Time Adaptation by Entropy Minimization},
  author    = {Wang, Dequan and Shelhamer, Evan and Liu, Shaoteng and Olshausen, Bruno and Darrell, Trevor},
  booktitle = {International Conference on Learning Representations},
  year      = {2021},
  url       = {https://openreview.net/forum?id=uXl3bZLkr3c}
}

@inproceedings{niu2022efficient,
  title={Efficient test-time model adaptation without forgetting},
  author={Niu, Shuaicheng and Wu, Jiaxiang and Zhang, Yifan and Chen, Yaofo and Zheng, Shijian and Zhao, Peilin and Tan, Mingkui},
  booktitle={International conference on machine learning},
  pages={16888--16905},
  year={2022},
  organization={PMLR}
}

@article{Things2019image2,
  title={THINGS: A database of 1,854 object concepts and more than 26,000 naturalistic object images},
  author={Hebart, Martin N and Dickter, Adam H and Kidder, Alexis and Kwok, Wan Y and Corriveau, Anna and Van Wicklin, Caitlin and Baker, Chris I},
  journal={PLOS ONE},
  volume={14},
  number={10},
  pages={e0223792},
  year={2019},
  publisher={Public Library of Science San Francisco, CA USA}
}

@inproceedings{spampinato2017deep,
  title     = {Deep Learning Human Mind for Automated Visual Classification},
  author    = {Spampinato, Concetto and Palazzo, Simone and Kavasidis, Isaak and Giordano, Daniele and Souly, Nasim and Shah, Mubarak},
  booktitle = {Proceedings of the IEEE Conference on Computer Vision and Pattern Recognition (CVPR)},
  pages     = {4503--4511},
  year      = {2017}
}

@inproceedings{albuquerque2019cross,
  title={Cross-subject statistical shift estimation for generalized electroencephalography-based mental workload assessment},
  author={Albuquerque, Isabela and Monteiro, Jo{\~a}o and Rosanne, Olivier and Tiwari, Abhishek and Gagnon, Jean-Fran{\c{c}}ois and Falk, Tiago H},
  booktitle={2019 IEEE International Conference on Systems, Man and Cybernetics (SMC)},
  pages={3647--3653},
  year={2019},
  organization={IEEE}
}

@article{ma2022large,
  title={A large EEG dataset for studying cross-session variability in motor imagery brain-computer interface},
  author={Ma, Jun and Yang, Banghua and Qiu, Wenzheng and Li, Yunzhe and Gao, Shouwei and Xia, Xinxing},
  journal={Scientific Data},
  volume={9},
  number={1},
  pages={531},
  year={2022},
  publisher={Nature Publishing Group UK London}
}

@article{radovanovic2010hubs,
  title={Hubs in space: Popular nearest neighbors in high-dimensional data},
  author={Radovanovic, Milos and Nanopoulos, Alexandros and Ivanovic, Mirjana},
  journal={Journal of Machine Learning Research},
  volume={11},
  number={sept},
  pages={2487--2531},
  year={2010}
}

@article{dinu2015hubness,
  title={Improving zero-shot learning by mitigating the hubness problem},
  author={Dinu, Georgiana and Lazaridou, Angeliki and Baroni, Marco},
  journal={arXiv preprint arXiv:1412.6568},
  year={2014}
}

@inproceedings{lazaridou2015hubness,
  title={Hubness and pollution: Delving into cross-space mapping for zero-shot learning},
  author={Lazaridou, Angeliki and Dinu, Georgiana and Baroni, Marco},
  booktitle={Proceedings of the 53rd Annual Meeting of the Association for Computational Linguistics and the 7th International Joint Conference on Natural Language Processing (Volume 1: Long Papers)},
  pages={270--280},
  year={2015}
}

@inproceedings{guo2017calibration,
  title={On calibration of modern neural networks},
  author={Guo, Chuan and Pleiss, Geoff and Sun, Yu and Weinberger, Kilian Q},
  booktitle={International conference on machine learning},
  pages={1321--1330},
  year={2017},
  organization={PMLR}
}

@inproceedings{shigeto2015ridge,
  title={Ridge regression, hubness, and zero-shot learning},
  author={Shigeto, Yutaro and Suzuki, Ikumi and Hara, Kazuo and Shimbo, Masashi and Matsumoto, Yuji},
  booktitle={Joint European conference on machine learning and knowledge discovery in databases},
  pages={135--151},
  year={2015},
  organization={Springer}
}

@InProceedings{radford2021clip,
  title = 	 {Learning Transferable Visual Models From Natural Language Supervision},
  author =       {Radford, Alec and Kim, Jong Wook and Hallacy, Chris and Ramesh, Aditya and Goh, Gabriel and Agarwal, Sandhini and Sastry, Girish and Askell, Amanda and Mishkin, Pamela and Clark, Jack and Krueger, Gretchen and Sutskever, Ilya},
  booktitle = 	 {Proceedings of the 38th International Conference on Machine Learning},
  pages = 	 {8748--8763},
  year = 	 {2021},
  editor = 	 {Meila, Marina and Zhang, Tong},
  volume = 	 {139},
  series = 	 {Proceedings of Machine Learning Research},
  month = 	 {18--24 Jul},
  publisher =    {PMLR},
}

@inproceedings{sun2020ttt,
  title={Test-time training with self-supervision for generalization under distribution shifts},
  author={Sun, Yu and Wang, Xiaolong and Liu, Zhuang and Miller, John and Efros, Alexei and Hardt, Moritz},
  booktitle={International conference on machine learning},
  pages={9229--9248},
  year={2020},
  organization={PMLR}
}

@inproceedings{zhang2022memo,
 author = {Zhang, Marvin and Levine, Sergey and Finn, Chelsea},
 booktitle = {Advances in Neural Information Processing Systems},
 editor = {S. Koyejo and S. Mohamed and A. Agarwal and D. Belgrave and K. Cho and A. Oh},
 pages = {38629--38642},
 publisher = {Curran Associates, Inc.},
 title = {MEMO: Test Time Robustness via Adaptation and Augmentation},
 url = {https://proceedings.neurips.cc/paper_files/paper/2022/file/fc28053a08f59fccb48b11f2e31e81c7-Paper-Conference.pdf},
 volume = {35},
 year = {2022}
}

@inproceedings{smith2017invertedsoftmax,
  title     = {Offline Bilingual Word Vectors, Orthogonal Transformations and the Inverted Softmax},
  author    = {Smith, Samuel L. and Turban, David H. P. and Hamblin, Steven and Hammerla, Nils Y.},
  booktitle = {International Conference on Learning Representations},
  year      = {2017},
  url       = {https://openreview.net/forum?id=r1Aab85gg}
}

@inproceedings{iscen2017diffusion,
  title={Efficient diffusion on region manifolds: Recovering small objects with compact cnn representations},
  author={Iscen, Ahmet and Tolias, Giorgos and Avrithis, Yannis and Furon, Teddy and Chum, Ondrej},
  booktitle={Proceedings of the IEEE conference on computer vision and pattern recognition},
  pages={2077--2086},
  year={2017}
}
}


\end{document}